\RequirePackage{amsmath}
\documentclass[runningheads]{llncs}
\usepackage[T1]{fontenc}
\usepackage{graphicx}
\usepackage[utf8]{inputenc}
\usepackage{amsmath}
\usepackage{amssymb}
\usepackage[boxed]{algorithm2e}
\usepackage{placeins}
\usepackage{soul}
\usepackage{xcolor}

\newcommand{\Bem}[1]{}

\newcommand{\one}{\mathbf{1}}
\newcommand{\diag}{\textrm{diag}}

\usepackage{amsmath}

\begin{document}
\title{A Method for Handling Negative Similarities in Explainable Graph Spectral Clustering of Text Documents - Extended Version \thanks{Supported by Polish Ministry of Science}}
\titlerunning{Handling Negative Similarities in XGSC}

\author{Mieczys{\l}aw A. K{\l}opotek\inst{1}
\orcidID{0000-0003-4685-7045} 
\and \\
S{\l}awomir T.  Wierzcho{\'n}\inst{1}\orcidID{0000-0001-8860-392X}
\and\\
Bart{\l}omiej Starosta \inst{1}\orcidID{0000-0002-5554-4596}
 \and \\ Dariusz Czerski \inst{1}\orcidID{0000-0002-3013-3483}
\and \\ Piotr Borkowski \inst{1}\orcidID{0000-0001-9188-5147}}
\authorrunning{M. K{\l}opotek et al.}

\institute{Institute of Computer Science, Polish Academy of Sciences, ul. Jana Kazimierza 5, 01-248 Warsaw, Poland\\
\email{\{klopotek,barstar,stw,dcz,pbr\}@ipipan.waw.pl}, 
\url{http://www.ipipan.waw.pl} 
}
\maketitle              
\begin{abstract}
This paper investigates the problem of Graph Spectral Clustering with negative similarities,  
resulting from document embeddings different from the traditional Term Vector Space (like doc2vec, GloVe, etc.). 
Solutions for combinatorial Laplacians and normalized Laplacians are discussed. An experimental investigation shows the advantages and disadvantages of 6 different solutions proposed in the literature and in this research.
{The research demonstrates that GloVe embeddings frequently cause failures of normalized Laplacian based GSC due to negative similarities. Furthermore, 
application of methods curing similarity negativity leads to  accuracy improvement for both combinatorial and normalized Laplacian based GSC. 
It also leads to applicability for GloVe embeddings of explanation methods developed originally bythe authors for Term Vector Space embeddings.\footnote{This is an extended version of a paper accepted for the 
25th International Conference on Computational Science (ICCS), 7-9 July 2025.} 
}

\keywords{
Artificial Intelligence \and
Machine Learning \and 
Graph spectral clustering  \and 
document embeddings \and 
negative similarities
}
\end{abstract}
%

\section{Introduction}

Graph Spectral Clustering (GSC) is a frequently used and highly effective method for clustering textual documents. 
{It approximates complex graph clustering methods based on RCut and NCut criteria well, while maintaining low computational complexity.}
GSC is also highly competitive compared to the methods of clustering documents in the Term Vector Space (TVS) because GSC clusters in a space of several dimensions, while clustering in TVS engages at least thousands of dimensions. 
GSC derives clusters from a matrix of similarities, which in the case of textual documents are traditionally cosine similarities of documents in TVS. 
%
%
GSC methods proceed as follows: A Laplacian is derived from the similarity matrix (e.g. combinatorial or normalized one). Then its eigen-decomposition is obtained. If the clustering splits the documents into $k$ clusters, $k$ eigenvectors corresponding to $k$ lowest eigenvalues are selected as an embedding for the documents and the clustering is performed, e.g., by the $k$-means algorithm in the space spanned by these eigenvectors. The method has fine mathematical properties in TVS as the similarities are non-negative and hence the mentioned Laplacians are positive semi-definite and eigen-decomposition yields real-valued eigenvectors and non-negative eigenvalues. 

Although TVS was used as a document embedding method, new methods such as
Word2vec \cite{Rong:2014:word2vec}, Doc2Vec \cite{lau2016empiricaldoc2vec}, GloVe \cite{Pennington:2014:glove}, BERT \cite{devlin2019bert}
based on transformers and others \cite{Lin:2022:transformerssurvey}.
{have emerged.} 
{These methods embed documents} in much lower dimensional space than TVS and reflect not only the set of words in the document but also semantic similarities between the words themselves. 
GSC can also be theoretically used in such embeddings, reducing dimensionality and taking advantage of cosine similarities. 
However, some problems arise when using these new embeddings. Unlike TVS embeddings, the (cosine) similarities between text documents may be negative. Of course, situations with negative similarities are well known, particularly in the context of analysis of social networks~\cite{Tang2016}. They have lived to see several constructive solutions, see e.g.~\cite{Hou2003}. Here positive and negative weights describe specific situations from the real world. But even so, as Knyazev \cite{Knyazev:2017} points out, there are many theoretical problems to handle.

{In this paper, we tackle the issue of clustering textual documents, concentrating on short texts (represented by tweets here). 
In our context, the problem is qualitatively different. Using, for example, the GloVe representation, the cosine similarity between two elements is a number that can be positive or negative. It is an abstract number with no physical meaning. Therefore, we face a serious technical problem when using GSC since combinatorial and normalized Laplacians are assumed to be positive semidefinite matrices only if the similarities are non-negative. 
{We also have problems with normalized Laplacians, as they can lead to complex numbers (square roots of diagonal elements of combinatorial Laplacians).}

The paper is organized as follows: 
Section \ref{sec:prev} reviews previous work on the related issues. Section \ref{sec:oursolution} contains our proposal for handling the issue of negative similarities, with subsection \ref{sec:combinatorial} concentrating on the combinatorial Laplacians, and subsection \ref{sec:normalized} concentrating on the normalized Laplacians. 
Section \ref{sec:experimental} presents the results of our experiments with the proposed methods. Section \ref{sec:conclusions} summarizes the paper. 

\section{Previous work}\label{sec:prev}

{}{
Graph Spectral Clustering is a methodology for low-complexity approximation of graph clustering based on graph cut criteria.
The best known criteria are RCut (ratio-cut, \cite{Leighton:1988}) and NCut (normalized cut, \cite{Hagen:1992}) described carefully, e.g., in \cite{STWMAKSpringer:2018} or \cite{Lux07}. 
}

\begin{align}
\label{eq:qRCut_def}
RCut(\Gamma) = \sum_{j=1}^k \frac{cut(C_j,\bar{C}_j)}{|C_j|}=\sum_{j=1}^k \frac{1}{|C_j|}\sum_{i \in C_j}\sum_{\ell \notin C_j}s_{i\ell} \\
\label{eq:qNCut_def}
NCut(\Gamma) = \sum_{j=1}^k \frac{cut(C_j,\bar{C}_j)}{\mathcal{V}_j}=\sum_{j=1}^k \frac{1}{\mathcal{V}_j}\sum_{i \in C_j}\sum_{\ell \notin C_j}s_{i\ell}
\end{align}
\noindent
{where} $s_{i\ell}$ is a similarity between objects $i$ and $\ell$ (usually $s_{i\ell} \in [0,1]$), $\Gamma$ is a partition of the set of objects, $|C_j|$ stands for the cardinality of cluster~$C_j$,
$    cut(C_j,\bar{C}_j) = \sum_{i \in C_j}\sum_{\ell \notin C_j} s_{i\ell}
$
is called Cut criterion \cite{Wu:1993} 
and $\mathcal{V}_j=\sum_{i}\sum_{\ell}s_{i\ell}$ is the volume of $j$-th cluster.
Note that in the above criteria, the object $i$ is meant to belong to cluster $C_j$, while the object $\ell$ does not. 
By convention $s_{ii}=0$ as acyclic graphs are handled by GSC.

A new criterion (NRCut) based on the two above can also be considered. 

\begin{align}
\label{eq:qNRCut_def}
NRCut(\Gamma)= \sum_{j=1}^k \frac{cut(C_j,\bar{C}_j)}{\mathcal{V}'_j}=\sum_{j=1}^k \frac{1}{\mathcal{V}'_j}\sum_{i \in C_j}\sum_{\ell \notin C_j}s_{i\ell}
\end{align}

We shall discuss here only two types of GSC. 
GSC with combinatorial Laplacian approximates RCut. 
GSC with normalized Laplacian approximates NCut. 
A \emph{combinatorial Laplacian} is defined as
\begin{equation}\label{eq:combLapDef} L=D-S, \end{equation}
\noindent
where $D$ is the diagonal matrix with $d_{ii}=\sum_{\ell=1}^ns_{i\ell}$ for each $i \in [n]$. 
A \emph{normalized Laplacian} $\mathcal{L}$ of the graph represented by~$S$ is defined as 
\begin{equation}\label{eq:normLapDef}\mathcal{L}=D^{-1/2}L D^{-1/2}= I -D^{-1/2}S D^{-1/2} .\end{equation}%
The \emph{rationormalized Laplacian}%
takes the form
\begin{align}\label{eq:ratnormLapDef}\mathcal{L_R}=&{D'}^{-1/2}L {D'}^{-1/2}= 
I -{D'}^{-1/2}S' {D'}^{-1/2} 
\end{align}%
\noindent
where $S'=S+I$ and $D'=D+I$ ($I$ - identity matrix).

There exist numerous application areas where it is convenient to use negative similarity measures.
They include, but are not limited to, studies based on correlations \cite{Knyazev:2015}, investigations of electric networks \cite{Zelazo:2014}, and others. 
As mentioned in the Introduction, such similarity measures constitute various problems both for the graph cut criteria and the GSC clustering methods if we want to extend them into such a realm.

To overcome the problems with negative similarities, several proposals were elaborated. 
One can eliminate negative similarities  setting them to zero\footnote{The proposal of signed cuts in \cite{Kunegis:2010} ignores in fact negative weights, see \cite{Knyazev:2017}. }

\begin{equation}\label{eq:sp_zeroneg}
s_{ik}^{(pZ)} = \left\{ \begin{array}{ll}
s_{ik} & \textrm{if $s_{ik}>0$}\\
0 & \textrm{otherwise}
\end{array}
\right.
\end{equation}
Other simple possibilities include taking absolute values, or adding a positive constant to all edge weights. Approaches depend on the application, i.e.   why some weights are negative and what the negativity means.

\begin{table}
\caption{TWT.10 datasets - hashtags and cardinalities of the set of related tweets  used in the experiments in each sample}\label{tab:twt10set}
\centering
{\tiny
\begin{tabular}{|r|l|c|}
\hline
& Sample 0 &\\
\hline
    No.  & hashtag & count \\
    \hline
  0& 90dayfiance & 316\\
	 1& tejran & 345\\
	 2& ukraine & 352\\
	 3& tejasswiprakash & 372\\
	 4& nowplaying & 439\\
	 5& anjisalvacion & 732\\
	 6& puredoctrinesofchrist & 831\\
	 7& 1 & 1105\\
	 8& lolinginlove & 1258\\
	 9& bbnaija & 1405\\
  \hline
\end{tabular}
\hspace{0.5cm}
\begin{tabular}{|r|l|c|}
\hline
& Sample 1 &\\
\hline
    No.  & hashtag & count \\
    \hline
 0 &  bb23  &  1723\\
	 1 &  lfc  &  1751\\
	 2 &  aewdynamite  &  1821\\
	 3 &  blm  &  1849\\
	 4 &  trump  &  1910\\
	 5 &  maga  &  2079\\
	 6 &  shinycheck  &  2235\\
	 7 &  nufc  &  2435\\
	 8 &  cdnpoli  &  2451\\
	 9 &  2  &  2772\\
  \hline
\end{tabular}
\hspace{0.5cm}
\begin{tabular}{|r|l|c|}
\hline
& Sample 2 &\\
\hline
    No.  & hashtag & count \\
    \hline
             0 &  rhobh  &  1298 \\
         1 &  robostopia  &  1323 \\
         2 &  gh  &  1398 \\
         3 &  lufc  &  1470 \\
         4 &  btc  &  1487 \\
         5 &  demdebate  &  1491 \\
         6 &  browns  &  1493 \\
         7 &  brexit  &  1607 \\
         8 &  bb22  &  1622 \\
         9 &  covid\_19  &  1696 \\
  \hline
\end{tabular}
\qquad
\begin{tabular}{|r|l|c|}
\hline
& Sample 3 &\\
\hline
    No.  & hashtag & count \\
    \hline
    	 0 &  s  &  1141 \\
	 1 &  sidnaaz  &  1153 \\
	 2 &  anitwt  &  1154 \\ 
	 3 &  breaking  &  1154 \\
	 4 &  rhop  &  1154 \\
	 5 &  treasure  &  1162 \\
	 6 &  cfc  &  1167 \\
	 7 &  trump2020  &  1222 \\
	 8 &  avfc  &  1278 \\
	 9 &  3  &  1293 \\
  \hline
\end{tabular}
\hspace{0.5cm}
\begin{tabular}{|r|l|c|}
\hline
& Sample 4 &\\
\hline
    No.  & hashtag & count \\
    \hline
     0 &  mentalhealth  &  1003             \\
	 1 &  bbcqt  &  1025             \\
	 2 &  vote  &  1025             \\
	 3 &  dnd  &  1031             \\
	 4 &  r4today  &  1036             \\
	 5 &  nffc  &  1036             \\
	 6 &  smackdown  &  1063             \\
	 7 &  debates2020  &  1065             \\
	 8 &  election2020  &  1093             \\
	 9 &  nfl  &  1122             \\
  \hline
\end{tabular}
} 
\end{table}

The mentioned difficulties are avoided by using the so-called \emph{signed Laplacian}. 
It differs from the traditional graph Laplacian in the computation of the $D$ matrix, where $d_{ii}$ is not the sum of $s_{ij}$, but rather of $|s_{ij}|$. This makes the signed Laplacian positive semi-definite
and its eigenvalues are non-negative, so that the definitions of the
Fiedler vector and the normalized Laplacian can be formally reused, see \cite{Gallier:2016}. 
Chen et al. 
\cite{Chen:2021} investigate signed Laplacian positive semi-definiteness in more detail for special cases.
However, Knyazev \cite{Knyazev:2017} highlights several weaknesses of signed Laplacians, when the results are to be interpreted, in particular with interpretation of the results in terms of mass-springs compared to the traditional Laplacian. 
Knyazev 
\cite{Knyazev:2015} discusses issues related to negative weights when creating signal filters. 
Zelazo and Buerger 
\cite{Zelazo:2014} investigate  conditions under which Laplacians for graphs with negative weights are positive semi-definite, formulating them in terms of an effective resistance model.  
\cite{Chen:2016} reiterates these results and provides some geometric interpretations.
\cite{Ahmadizadeh:2017} 
{formulates necessary and sufficient conditions for Laplacians to be positive semi-definite.} 
Grudsky et al. 
\cite{GRUDSKY:2024} investigate the problem of Laplacian eigenvalues in the case of graphs with a single negative weighted edge. 
Tian et al. 
\cite{TIAN:2022} discusses the way of handling negative weights in the context of multi-agent systems.

\section{Our approach to technical problems}\label{sec:oursolution}

Negative cosine similarities are an issue with modern document embeddings, as seen in Table \ref{tab:Data_X.prop} for Tweeter samples listed in Table~\ref{tab:twt10set}. For details, see Section \ref{sec:experimental}. 
%

\subsection{The problem of combinatorial Laplacians}\label{sec:combinatorial}

When discussing extensions of the GSC methodology to cases of negative similarities, we study various transformations of the similarity measure. 
 Essentially, we want the clustering that considers original similarities, and those after transformation shall not change. 
This is because if we consider any arbitrary partition into $k$ clusters
So, let us ask what will happen with a clustering fitting RCut, defined in the equation \eqref{eq:qRCut_def}, if we add a positive number, say $c$, to all non-diagonal similarities. In such a case, new matrix~$\tilde{S}$ takes the form
\begin{equation}\label{eq:dodawaniestalvejMacierz}
\tilde{S} = S + c(J - I)
\end{equation}
\noindent
where $I$ denotes the identity matrix, $J=\one \one^T$ is the matrix with all elements equal one, and $\one$ is the vector with all entries equal one. The degree matrix $\tilde{D}$ corresponding to~$\tilde{S}$ is
\begin{equation}
    \tilde{D} = \diag(\tilde{S}\one) = \diag(S\one + c(n-1)\one) = D + c(n-1)I     
\end{equation}
\noindent
where $\diag(v)$ returns a diagonal matrix with $v$ as its diagonal. Finally, the new Laplacian is
\begin{equation}
\tilde{L}= \tilde{D} - \tilde{S} = L - cJ + cnI
\end{equation}

\begin{table}
\caption{Properties of the dataset  en.Size=150.TagCap=300.SEL.10tags -- the number of negative similarities ($S$ matrix)  and the number of negative elements of the row sum matrix ($D$ matrix.) in each sample. }
\label{tab:Data_X.prop}
\begin{tabular}{|l|l|r|r|}
\hline
Sample & 
Embedding  & negative    & negative \\
 & 
type &   $S$ entries  & $D$ entries\\
\hline
0 & WikiGloVe   &  4256724 &   171   \\
0 & TweetGloVe    &  242824 & 0 \\
\hline
1 & WikiGloVe   &  147268 &   3   \\
1 & TweetGloVe    &  14 & 0 \\
\hline
2 & WikiGloVe   &  2032 &   0   \\
2 & TweetGloVe    &  0 & 0 \\
\hline  
\end{tabular} %
\hspace{0.6cm}
\begin{tabular}{|l|l|r|r|}
\hline
Sample & 
Embedding  & negative    & negative \\
 & 
type &   $S$ entries  & $D$ entries\\
\hline
3 & WikiGloVe   &  54 &   0   \\
3 & TweetGloVe    &  16 & 0 \\
\hline  
4 & WikiGloVe   & 52552 & 2     \\
4 & TweetGloVe    & 0  &  0 \\
\hline  
\end{tabular}
\end{table}

Let $(\lambda,v)$ be an eigenpair of the Laplacian~$L$. Then
\begin{equation}
    \tilde{L}v = Lv - cJv + cmIv = (\lambda+cn)v
\end{equation}
since $Jv=\mathbf{0}$. This shows that $(\lambda+cn,v)$ is an eigenpair of~$\tilde{L}$. So, RCut minimizing clustering remains unchanged under such an operation. 

Note that we can add the constant $c$ to all the elements of the similarity matrix obtaining $\tilde{S}=S+cJ$. Then $\tilde{D}=D+cnI$, and we conclude that if $(\lambda,v)$ is an eigenpair of $L$ then $(\lambda+cn,v)$ is an eigenpair of such modified Laplacian matrix. 

Alternatively, we can add a positive constant~$\alpha$ to the diagonal elements of the degree matrix, that is, $\hat{D}=D+\alpha I$. Then for any eigenpair of the Laplacian $L$
\begin{equation}
(\hat{D}-S)v = Lv + \alpha v=(\lambda+\alpha)v
\end{equation}
Although the matrix $(\hat{D}-S)$ does not fulfill the requirements of being a combinatorial Laplacian, it still belongs to a large family of generalized graph Laplacians,~\cite{Biy07}. Rocha and Trevisan call such matrices perturbed Laplacians and develop their theory in~\cite{Rocha:2016}.

Now consider its approximation with combinatorial Laplacian based GSC. 

To have similarities ranging from 0 to 1 (and not up to 1+c), one needs to divide both sides by (1+c). 
Note that the formula \eqref{eq:dodawaniestalvejMacierz} translates to off-diagonal elements of the matrix $\tilde{S} $
\begin{equation}\label{eq:sp_cadd}
    s_{ik}^{(pA)}= s_{ik}+c
\end{equation}
The similarities $s_{ik}^{(pA)}$ will get out of the range $[0,1]$ for large enough $c$. To get them again into this range, we can divide them by $c+1$, leading to elements of the matrix $\bar{S} = \frac{S + c(J - I)}{1+c}$ of the form: 
\begin{equation}\label{eq:sp_cnorm}
    s_{ik}^{(pN)}=\frac{s_{ik}+c}{1+c}
\end{equation}
which will lead to the same eigenvectors of the resulting combinatorial Laplacian $\bar{L}=\bar{D}-\bar{S}$ with $\bar{D}=diag(\bar{S}\one)$ as for the original $L$.   

Conclusion: The calculation of Laplacian $\tilde{L},\hat{L}, \bar{L}$ is not necessary, because the eigenvectors of the original $L$ will not differ. Hence also the clustering based on lowest eigenvectors will yield the same results.

\subsubsection{Geometric Interpretation of $s_{ik}^{(pN)}$}
Interestingly, the formula \eqref{eq:sp_cnorm} can be assigned a geometric interpretation if we compute the similarities as cosines between the document embedding vectors in an N-dimensional space, such as the doc2vec or GloVe space. 

\begin{figure}
\centering
\includegraphics[width=0.85\textwidth]{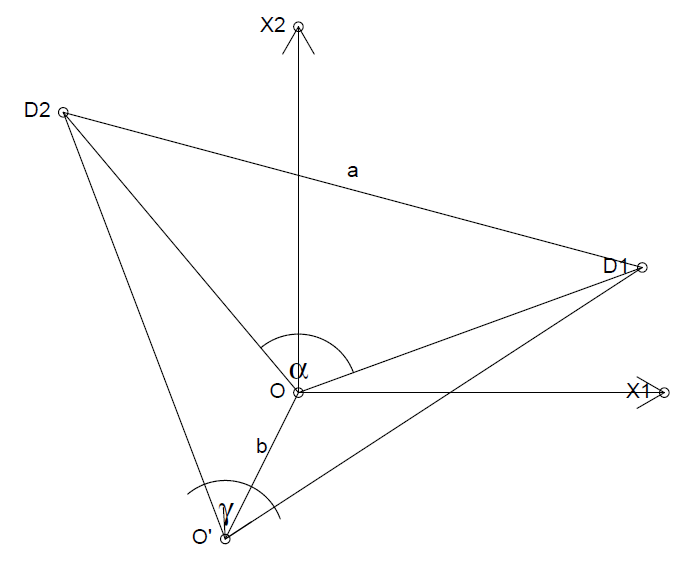} %
\caption{Geometric interpretation of removing negativity by adding a constant 
}\label{fig:negaddc}
\end{figure}

Assume that we extend this space by the N+1st dimension, orthogonal to all other dimensions, and move the documents by a vector of length $b$ along this dimension. 
In Figure \ref{fig:negaddc} it is illustrated as follows: Pick up two document embedding vectors, D1 and D2 (normalized to a unit length) and let the angle between them be $\alpha$ in the original space. 
Note that the three points D1, D2 and O (the origin of the coordinate system) span a hyperplane in the $N$ dimensional space, and this plane may be spanned by e.g. two mutually orthogonal unit vectors $\vec{OX1}, \vec{OX2}$. 
We add the $(N + 1)$st dimension and place the point $O'$ on the $(N + 1)$st axis at the distance $b$ from the origin. 
The vector $\vec{OO'}$ is orthogonal to all vectors in the N-dimensional space, including $\vec{OD1},\vec{OD2}$. 
Let the angle between the vectors $\vec{O'D1},\vec{O'D2}$ representing documents in a ``shifted space'' be 
 $\gamma`(b)$. Whereby (obviously) $\gamma(0)=\alpha$.

According to the extended version of Pitagoras Theorem 
$$
\cos(\alpha)=
\frac{|OD1|^2 +|OD2|^2-|D1D2|^2}{2 |OD1| |OD2|}
$$
and
$$
\cos(\gamma(b))=
\frac{|O`D1|^2 +|O`D2|^2-|D1D2|^2}{2 |O`D1| |O`D2|}
$$
Denote $|D1D2|$ with $a$.
Based on the assumption that $|OD1| =|OD2|=1$ we get:
\[
\begin{split}
\cos(\alpha) &= \frac{1^2+1^2-a^2}{2 \cdot 1 \cdot 1} \Rightarrow 2\cos(\alpha) = 2-a^2 \\
 & \Rightarrow a^2 = 2-2\cos(\alpha)
\end{split}
\] 
On the other hand: 
\[
\begin{split}
    \cos(\gamma(b)) &= \frac{b^2+1^2+b^2+1^2-a^2}{2 \cdot \sqrt{b^2+1^2}  \cdot \sqrt{b^2+1^2}} = 1-\frac{a^2}{2 \cdot (b^2+1^2)}\\
 &= 1-\frac{2-2\cos(\alpha)}{2 \cdot (b^2+1^2)} = 
1-\frac{1-\cos(\alpha)}{ (b^2+1^2)}\\
& = \frac{\cos(\alpha)+b^2}{ (1+b^2)}
\end{split}
\]
So $b^2$ is the constant $c$ from formula \eqref{eq:sp_cnorm}, and otherwise the formulas are identical.

\subsection{The problem of normalized Laplacians}\label{sec:normalized}

Two types of problems with computation of normalized Laplacian  $\mathcal{L}=D(S)^{-1/2}LD(S)^{-1/2}$ may occur.

First, when $L$ contains positive off-diagonal elements while all diagonal elements are positive, so that $\mathcal{L}$ is computable, but some off-line elements remain positive.   
Curing this situation is analogous to combinatorial Laplacian and shall not be detailed here. 

Second type of problem, as described among others in \cite{Knyazev:2017}, when some elements of $D$ are negative. 

Square rooting of a negative number on the diagonal $D(S)$ would result in a complex number and break the basic assumptions of GSC based on normalized Laplacians. 
But the problem is more profound. 
This GSC approximates clustering based on the minimization of NCut criterion. This criterion refers to the cluster volume that may turn out to be negative. A cluster with negative volume -- that is with strongly dissimilar documents --  has a chance to minimize the NCut criterion -- instead of clusters with strongly similar documents one gets ones with strongly dissimilar ones. This issue strongly resembles the problems with kernel $k$-means which may not reach the minimum of $k$-means criterion. 
Therefore, the NCut criterion must be addressed at the very beginning. 

Let us consider how serious is the risk of negative elements $D$. 
If all elements of $D$ were positive, the sum of all elements of $S$ would also be positive.
So consider a bit abstract embedding into one-dimensional space. Then normalized document embedding vectors would be either [1] or [-1]. Let $p$ documents have the coordinate [1] and $q=n-p$ documents have the coordinate [-1]. 
In such a case $p(p-1)+q(q-1)$ ordered pairs of documents would have similarity 1, while $2pq$ would be of similarity -1. So, the sum of $S$ matrix would be
$p(p-1)+q(q-1)-2pq=(p-q)^2-p-q=(p-q)^2-n$. In the worst case, the sum of $S$ would be $-n$. In higher dimensions, this would be less serious, but still, the risk of negativity remains, independently of the size of the document collection. 

We will consider several proposals. 
\begin{itemize}
    \item Adding a constant to diagonal of combinatorial Laplacian,
    \item Adding a constant to each similarity of combinatorial Laplacian,
    \item Manipulating similarity computation by taking not the cosine of the angle between documents, but half of this angle.
    \item Replacing similarity with the exponent of the negated distance between documents on a unit sphere. 
\end{itemize}

So, let us ask what will happen if we add a positive number to all similarities. Essentially, the clustering taking really similarities (and not their substitutes) into account shall not change. This is what is meant by RCut. 
Normalized Laplacian will be computable, but does it make sense?
Will the NCut change under such an operation?

Let us investigate some details of the latter proposal. 

 Define 
 \begin{equation} s_{ik}^{(pD)}(x)=\frac{s_{ik}}{\sqrt{d_{ii}+x}\sqrt{d_{kk}+x}}\end{equation}
 Note that $s_{ik}^{(pD)}(0)$ is the negated off-diagonal element of normalized Laplacian. 

It can be shown that if $s_{ik}^{(pD)}(0)>s_{i\ell}^{(pD)}(0)$, then 
$s_{ik}^{(pD)}(c)>s_{i\ell}^{(pD)}(c)$, for $c>0$ and  $s_{ik}>s_{i\ell}$. 
Consider three documents, $i,k,\ell$ and let 
$$s_{ik}^{(pD)}(c)>s_{i\ell}^{(pD)}(c)$$
which means
$$\frac{s_{ik}}{\sqrt{d_{ii}+c}\sqrt{d_{kk}+c}}
>
\frac{s_{i\ell}}{\sqrt{d_{ii}+c}\sqrt{d_{\ell\ell}+c}}
$$

$$\frac{s_{ik}}
{\sqrt{d_{kk}+c}}
>
\frac{s_{i\ell}}{\sqrt{d_{\ell\ell}+c}}
$$
If we are in the realm of non-negative similarities (other cases can be handled similarly)

$$\frac{s_{ik}}
{s_{i\ell}}
>
\frac{\sqrt{d_{kk}+c}}{\sqrt{d_{\ell\ell}+c}}
$$

$$\frac{s_{ik}^2}
{s_{i\ell}^2}
>
\frac{d_{kk}+c}{d_{\ell\ell}+c}
$$
With an increase of $c$, the expression on the right-hand side grows/decreases towards one. 
Therefore, if originally $s_{ik}>s_{i\ell}$, then the expression is true. 

This means that adding a constant to the normalized Laplacian diagonal keeps to a great extent the ordering of similarities, so that the results of clustering may be similar, unless the normalization changes proportions between similarities in the original Laplacian.
In case of some negative $d_{ii}$, adding an appropriate constant may turn the Laplacian into a computable one, resulting in clustering similar to the one originally intended. 

However, the problem is that this solution tends to be in fact a version of the newly introduced NRCut \cite{MonoSiedlce:2025}, and not NCut.
So another approach is needed. 

Another solution would be adding a constant $c$ to each similarity ($s^{(pA)}_{ik}(c)=s_{ik}+c)$.
Again, no warranty that the ordering of all normalized similarities will be the same and hence that clustering result is the same.

One solution could be to transform the similarity matrix $S$ into a positive one $S^{(pQ)}$ as follows:
\begin{equation} \label{eq:sp_cos2}
s_{ik}^{(pQ)}= \cos \left(\frac \pi 2 \frac{\arccos s_{ik}}{\max_{{i,k \in [n],i\ne k}} \arccos s_{ik} } \right)\end{equation}
whereby $\max$ is computed over all off-diagonal elements of the $S$ matrix.  
{$\cos$} {is non-negative in the range} {$[0,\frac{\pi}{2}]$} {while it is negative for greater angles. By dividing the actual angles between documents by the maximal angle, and multiplying with} {$\frac{\pi}{2}$} {we scale all the angles into the non-negative  cosine range.} 
Now, the traditional normalized Laplacian is applicable. 
The ranking of similarities of combinatorial Laplacian is preserved completely, but again no warranty for the normalized similarities.   {The above formula can be generalized to:}
\begin{equation} \label{eq:sp_cos2c}
s_{ik}^{(pC)}= \cos \left(\frac{\arccos s_{ik}}{1+c } \right)\end{equation}
{$c=1$} { is for sure a reasonable choice, because \emph{arccos} returns values in the range  } {$[0,\pi]$}  {and dividing this result by 2 scales them into the required range } {$[0,\frac{\pi}{2}]$}. 

{Note that if the graph has isolated nodes,}
we already get into trouble with normalized Laplacian because of division by zero; the same applies to the NCut criterion. 
{Therefore, a change in understanding NCut is needed in such a way that a cluster with all nodes isolated has a non-zero volume.  }
So, the similarity needs to be transformed. As the Euclidean distance between two normalized vectors $x_i,x_j$ equals to $\|x_i-x_j\|^2 = 2(1-s_{ij})$, where $s_{ij}=cos(x_i,x_j)$, our proposal is 
\begin{equation}s_{ik}^{(pE)}=e^{-(1-s_{ik})/2}
\end{equation}
This should be applied to get a new similarity matrix $S'$ as well as a redefinition of NCut to NCut$^{(pE)}$ (based on the new similarities). There is no need to worry about isolated nodes.
{Note that if we generalize the transformation} {$s_{ik}^{(pE)}$} to 
\begin{equation}
\label{eq:exp_c}
s_{ik}^{(pE)}(c)=e^{-(1-(s_{ik}+c))/2}
\end{equation}
{the normalized Laplacian will remain the same for all values}  
{$c\ge 0$} {because adding a constant in the exponent is the same as multiplying the similarity with another constant. } 

\subsubsection{Transformation preserving some properties of normalized Laplacian}
For combinatorial Laplacians, We mentioned a  solution 
\eqref{eq:sp_cnorm}
$   s_{ik}^{(pN)}=\frac{s_{ik}+c}{1+c}$
 consisting in adding a constant $c$ to each similarity and then normalizing. While with normalized Laplacian, we do not have a warranty that the ordering of all normalized similarities will be the same and hence that clustering result is the same, this ordering can be preserved if the similarity matrix has reasonable properties mentioned below,  
so that related properties of $D^{-0.5}SD^{-0.5}$ will be preserved and the clustering shall not deviate significantly.

Let us investigate some details of the transformation $s_{ik}^{(pN)}$. 
Assume 
$s_{ik}$, as a product of unit document vectors $\mathbf{v}_i, \mathbf{v}_j$, range from 0 to 1. 
Add a new dimension and make new unit document vector:
$\mathbf{v}_i'=(\mathbf{v}^T_i,\sqrt{x})^T/\sqrt{1+x}$. 
New $s_{ik}'=\frac{s_{ik}+x}{1+x}$. 
New $d_{ii}'=\sum_k s_{ik}'=\frac{d_{ii}}{1+x}+\frac{(n-1)x}{1+x}$

 Define as a consequence the "normalized"  $s_{ik}^{(v)}$: 
 \begin{equation} s_{ik}^{(v)}
= \frac{s_{ik}'}{\sqrt{d_{ii}'}\sqrt{d_{kk}'}}
= 
\frac{\frac{s_{ik}+x}{1+x}}
{
\sqrt{\frac{d_{ii}}{1+x}+\frac{(n-1)x}{1+x}}
\sqrt{\frac{d_{kk}}{1+x}+\frac{(n-1)x}{1+x}}
}
 \end{equation}
 that is, by multiplication by $1+x$  the nominator and the denominator of the right-hand-side.  
\begin{equation} 
s_{ik}^{(v)}
= \frac{s_{ik}+x}{\sqrt{d_{ii}+(n-1)x}\sqrt{d_{kk}+(n-1)x}}
 \end{equation}
 
Consider three documents, $i,k,\ell$.
Let $i,k$ belong to the same cluster, and $\ell$ to the other cluster. It would be reasonable to assume therefore that the similarity between $i$ and $K$ is bigger than between $i$ and $\ell$, that is 
$$s_{ik}>s_{i\ell}$$
Furthermore, the similarity between $i,k$ is reasonably bigger than average similarity between $k$ and anything else. 
Also, the similarity between $i,\ell$ is reasonably smaller than average similarity between $\ell$ and anything else. 

\textbf{Let us demonstrate} that under these reasonable assumptions
\begin{equation} \label{eq:toProve}
s_{ik}^{(v)} > s_{i\ell}^{(v)}
\end{equation}
for $x\ge 0$. 
The above expression is equivalent to:
\begin{equation} 
 \frac{s_{ik}+x}{\sqrt{d_{ii}+(n-1)x}\sqrt{d_{kk}+(n-1)x}}
>
 \frac{s_{i\ell}+x}{\sqrt{d_{ii}+(n-1)x}\sqrt{d_{\ell\ell}+(n-1)x}}
 \end{equation}

The above expression is equivalent to:
\begin{equation} 
 \frac{s_{ik}+x}{\sqrt{d_{kk}+(n-1)x}}
>
 \frac{s_{i\ell}+x}{\sqrt{d_{\ell\ell}+(n-1)x}}
 \end{equation}
Let us divide the denominator of both sides by $\sqrt{n-1}$
\begin{equation} 
 \frac{s_{ik}+x}{\sqrt{\frac{d_{kk}}{n-1}+x}}
>
 \frac{s_{i\ell}+x}{\sqrt{\frac{d_{\ell\ell}}{n-1}+x}}
 \end{equation}
As we assumed already, $\frac{d_{kk}}{n-1}<s_{ik}$ that is 
$\frac{d_{kk}}{n-1}=s_{ik}-\delta_k$. 
Also $\frac{d_{\ell\ell}}{n-1}>s_{i\ell}$ that is 
$\frac{d_{\ell\ell}}{n-1}=s_{i\ell}+\delta_\ell$. Whereby $\delta_k>0, \delta_\ell>0$.

Hence the above is equivalent:
\begin{equation} 
 \frac{s_{ik}+x}  {\sqrt{s_{ik}-\delta_k+x}}
 >
\frac{s_{i\ell}+x}{\sqrt{s_{i\ell}+\delta_\ell+x}}
 \end{equation}
Divide the nominator and denominator of the left hand side by $s_{ik}+x$, and the nominator and denominator of the 
right hand side by $s_{i\ell}+x$
\begin{equation} 
 \frac{1}  {\sqrt{\frac{1}{s_{ik}+x}-\frac{\delta_k}{(s_{ik}+x)^2}}}
 >
\frac{1}{\sqrt{\frac{1}{s_{i\ell}+x}+\frac{\delta_\ell}{(s_{i\ell}+x)^2}}}
 \end{equation}

\begin{equation} 
\sqrt{\frac{1}{s_{i\ell}+x}+\frac{\delta_\ell}{(s_{i\ell}+x)^2}}
>
 \sqrt{\frac{1}{s_{ik}+x}-\frac{\delta_k}{(s_{ik}+x)^2}}
 \end{equation}

\begin{equation} 
\frac{1}{s_{i\ell}+x}+\frac{\delta_\ell}{(s_{i\ell}+x)^2}
>
 \frac{1}{s_{ik}+x}-\frac{\delta_k}{(s_{ik}+x)^2}
 \end{equation}

\begin{equation} 
\frac{\delta_k}{(s_{ik}+x)^2}+
\frac{\delta_\ell}{(s_{i\ell}+x)^2}
>
 \frac{1}{s_{ik}+x}-\frac{1}{s_{i\ell}+x}
 \end{equation}

\begin{equation} 
\frac{\delta_k}{(s_{ik}+x)^2}+
\frac{\delta_\ell}{(s_{i\ell}+x)^2}
>
 \frac{(s_{i\ell}+x)-(s_{ik}+x)}{(s_{ik}+x)(s_{i\ell}+x)}
 \end{equation}

\begin{equation} \label{eq:final}
\frac{\delta_k}{(s_{ik}+x)^2}+
\frac{\delta_\ell}{(s_{i\ell}+x)^2}
>
 \frac{s_{i\ell}-s_{ik}}{(s_{ik}+x)(s_{i\ell}+x)}
 \end{equation}
As $s_{i\ell}<s_{ik}$, the righthandside is smaller than zero.  
As 
\begin{equation} 
\frac{\delta_k}{(s_{ik}+x)^2}+
\frac{\delta_\ell}{(s_{i\ell}+x)^2}
>0
 \end{equation}
and 
 \begin{equation}
0> \frac{s_{i\ell}-s_{ik}}{(s_{ik}+x)(s_{i\ell}+x)}
 \end{equation}
we see that \eqref{eq:final} is true, hence also \eqref{eq:toProve}.

\begin{table}
\caption{Clustering results of the dataset en.Size=150.TagCap=300.SEL.10tags sample 0 after similarity correction using formula \eqref{eq:sp_zeroneg} (setting negative $s_{ik}$ to zero)  -- F-score, averaged over 30 runs when normalized (left) and combinatorial (right) Laplacian based GSC was applied.
}
\label{tab:Data_0.N.Exp_0}
\begin{tabular}{|l|r|r|}
\multicolumn{3}{c}{Normalized Laplacian }\\
\hline
Embedding   & avg.   & SD of   \\
 type &  F-score       &   F-score \\
\hline
CountVectorizer    & 0.048 & 0.035  \\
\hline
TfVectorizer          & 0.037 & 0.019  \\
\hline
TfidfVectorizer        & 0.025 & 0.007  \\
\hline
WikiGloVe       &0.045 & 0.022  \\
\hline
TweetGloVe      & 0.060 & 0.036  \\
\hline  
\end{tabular}
\begin{tabular}{|l|r|r|}
\multicolumn{3}{c}{Combinatorial Laplacian }\\
\hline
Embedding   & avg.   & SD of   \\
 type &  F-score       &   F-score \\
\hline
CountVectorizer    & 0.0189 & 0.000  \\
\hline
TfVectorizer          & 0.019 & 0.000  \\
\hline
TfidfVectorizer        & 0.022& 0.001  \\
\hline
WikiGloVe       &0.020 &0.001  \\
\hline
TweetGloVe      & 0.018 & 0.000  \\
\hline  
\end{tabular}
\end{table}
\FloatBarrier

\subsection{Negativity versus Explainability}

{
GSC result explanation procedure elaborated in \cite{Plosone2025} encounters serious problems as it is based on the products of word embedding vectors and cluster center vectors which would lead to meaningless negative word importance. The correction proposed for combinatorial Laplacian based GSC keeps the spirit of \cite{Plosone2025}.  
As normalized Laplacian is concerned, we show in \cite{Plosone2025}  that additive corrections of similarity measure does not disturb the explanation bridge (its intermediate stage).
}

{
As combinatorial Laplacian is concerned,  
see in particular the formula (25) in \cite{Plosone2025}. Adding a new dimension, with constant coordinate for each word $\mathbf{w}$, as proposed in subsection \ref{sec:combinatorial},   
will not change anything in this formula and subsequent ones in subsection 5.3. of \cite{Plosone2025}. Hence the process of explanation would not change for TVS. As GloVe embedding would not differ in a formal way, the same applies also for GloVe embedding. So the explainability is granted. 
}

{
As normalized Laplacians are concerned, a quick review of subsection 6.3 of \cite{Plosone2025} indicates that the $\mathbf{g}$ vectors mentioned there will more strongly converge to 0 so that they are of smaller importance with the addition of a new dimension to the embedding with a constant coordinate does not disturb significantly the explanation process so that again the explainability is given. 
}

\section{Experiments}\label{sec:experimental}

We have performed experiments on the effectivity of GSC methods predicting hashtags for a large body of tweets, dataset  en.Size=150.TagCap=300.SEL.10tags,  using various methods of handling negativity, mentioned in the formulas 
\eqref{eq:sp_cadd},
\eqref{eq:sp_cnorm},
\eqref{eq:sp_cos2},
\eqref{eq:sp_cos2c},
\eqref{eq:exp_c}, \eqref{eq:sp_zeroneg}
for  $c=0,1,2,3$. 
$c=0$ represents the unadjusted case, whereas the other values represent adjustments to get rid of negative similarities. 
For the modified similarity matrices, both combinatorial and normalized Laplacians were used in GSC. 
The computations were performed for the traditional Term Vector Space (TVS, tf, tfidf) as well as for the GloVe based embeddings: TweetGlove (trained on Twitter data) and WikiGlove (trained on Wikipedia Data). 

The clustering experiments were performed with popular Python libraries: 
\texttt{numpy} \cite{NumPy:2020}, \texttt{scipy} \cite{SciPy:2020}, \texttt{scikit-learn} \cite{sklearnAPI:2013} and 
\texttt{soyclustering} \cite{soyclustering:020} which is an implementation of spherical $k$-means \cite{SKmeans:2020:113288}.
In particular, we used  
    \texttt{SpectralClustering} class from scikit-learn with two distinct  settings of the \texttt{affinity} parameter: \texttt{precomputed} (affinity from similarity matrix)   as a representative of the $L$-embedding and $\mathcal{L}$-embedding based clustering. 

    \subsection{Data Set Description}

For the initial experiment, we used a set, named en.Size=150.TagCap=300.SEL.10tags, consisting of random tweets published on Twitter (now X) between 2019 and 2023, with a length of more than 150 characters each, which are a collection of tweets related to the hashtags listed in Table~\ref{tab:twt10set}.    

Table \ref{tab:Data_X.prop} characterizes the dataset 
en.Size=150.TagCap=300.SEL.10tags
with respect to the two GloVe-based embeddings for the five considered samples. 
Obviously, the Term Vector Space embeddings  
{have} 
no negative similarity problems, so they were not included. 
TweetWiki embedding leads to numerous negative similarity matrix entries, but no problem with row sums occurs for our samples. 
The most difficult problems occur for the WikiGloVe embedding, as there are many more negative similarities and there are multiple rows with negative entries in three of the samples.   

\subsection{Results}
\newcommand{\naszLink}{\mbox{\url{https://github.com/ipipan-barstar/ICCS25.MfHNSiEGSCoTD}}}

Subsequent tables present clustering results for sample 0. Results for other samples can be accessed at the link \naszLink.

Let us first consider normalized Laplacians. 
Table \ref{tab:Data_0.N.Exp_0} (left) presents clustering results when the correction of negative similarities is based on zeroing them. 
Normalized Laplacian based clustering could be executed.
We see that modern (GloVe based)  do not have a big advantage over traditional (TVS based) embeddings. The results are the worst compared to other methods. 

\begin{table}
\caption{ Clustering results of the dataset en.Size=150.TagCap=300.SEL.10tags sample 0 after similarity correction 
using formula \eqref{eq:sp_cadd} (adding $c$ to all off-diagonal similarities, left) and 
using formula \eqref{eq:sp_cnorm} (adding $c$ and dividing by $1+c$ to all off-diagonal similarities, right) 
-- F-score, averaged over 30 runs when normalized Laplacian based GSC was applied.  
}
\label{tab:Data_0.N.Exp_2}
\label{tab:Data_0.N.Exp_1}
\begin{tabular}{|l|c|r|r|}
\multicolumn{3}{c}{using formula \eqref{eq:sp_cadd}}\\
\hline
Embedding   &$c$& avg.   & SD of   \\
 type & &  F-score       &   F-score \\
\hline
CountVectorizer    &0  & 0.039 & 0.022  \\
&1&0.083&0.029\\
&2&0.095&0.044\\
&3&0.091&0.042\\
\hline
TfVectorizer         &0& 0.038&0.026\\
&1&0.092&0.052\\
&2&0.093&0.041\\
&3&0.093&0.034\\
\hline
TfidfVectorizer      &0&0.026&0.009\\
&1&0.089&0.058\\
&2&0.087&0.054\\
&3&0.101&0.057\\
\hline
\hline
WikiGloVe   &0&  --- & ---  \\
&1&0.070&0.026\\
&2&0.093&0.050\\
&3&0.088&0.051\\
\hline
TweetGloVe    &0& 0.064&0.036\\
&1&0.081&0.047\\
&2&0.099&0.059\\
&3&0.084&0.060\\
\hline  
\end{tabular}
\begin{tabular}{|l|c|r|r|}
\multicolumn{3}{c}{using formula \eqref{eq:sp_cnorm}}\\
\hline
Embedding   &$c$& avg.   & SD of   \\
 type & &  F-score       &   F-score \\
\hline
CountVectorizer      &0  & 0.039 & 0.022  \\
&1&0.090&0.036\\
&2&0.091&0.044\\
&3&0.098&0.038\\
\hline
TfVectorizer           &0& 0.038&0.026\\
&1&0.088&0.040\\
&2&0.080&0.028\\
&3&0.088&0.044\\
\hline
TfidfVectorizer        &0&0.026&0.009\\
&1&0.082&0.042\\
&2&0.086&0.058\\
&3&0.096&0.055\\
\hline
\hline
WikiGloVe     &0&  --- & ---  \\
&1&0.094&0.039\\
&2&0.079&0.044\\
&3&0.084&0.051\\

\hline
TweetGloVe       &0& 0.064&0.036\\
&1&0.083&0.060\\
&2&0.085&0.064\\
&3&0.088&0.058\\
\hline  
\end{tabular}
\end{table}

\begin{table}
\caption{Clustering results of the dataset en.Size=150.TagCap=300.SEL.10tags sample 0 after similarity correction using formula \eqref{eq:sp_cos2}  (transformation $s_{ik} = cos( (\pi/2)(\arccos(s_{ik}) / \max \arccos(s_{ik})))$)  -- F-score, averaged over 30 runs when normalized (left) and combinatorial (right) Laplacian based GSC was applied.
}
\label{tab:Data_0.N.Exp_3}
\begin{tabular}{|l|r|r|}
\multicolumn{3}{c}{Normalized Laplacian }\\
\hline
Embedding   & avg.   & SD of   \\
 type &   F-score       &   F-score \\
\hline
CountVectorizer    & 0.041 & 0.022  \\
\hline
TfVectorizer          & 0.040 & 0.021  \\
\hline
TfidfVectorizer        & 0.024 & 0.007  \\
\hline
WikiGloVe       & 0.086 & 0.038  \\
\hline
TweetGloVe      & 0.070 & 0.041  \\
\hline  
\end{tabular}
\begin{tabular}{|l|r|r|}
\multicolumn{3}{c}{Combinatorial Laplacian }\\
\hline
Embedding   & avg.   & SD of   \\
 type &   F-score       &   F-score \\
\hline
CountVectorizer    &  0.019 & 0.000    \\
\hline
TfVectorizer          &  0.019 & 0.000  \\
\hline
TfidfVectorizer        &  0.022 & 0.001 \\
\hline
WikiGloVe       &   0.019 & 0.000 \\
\hline
TweetGloVe      &  0.018 & 0.000\\
\hline  
\end{tabular}
\end{table}

\begin{table}
\caption{Clustering results of the dataset en.Size=150.TagCap=300.SEL.10tags sample 0 
after similarity correction using formula \eqref{eq:sp_cos2c}  (transformation $s_{ik} = cos(arccos s_{ik} / (1+c)) $) (left) and 
after similarity correction using formula \eqref{eq:exp_c} (transformation $s_{ik} = \exp(-(1-(s_{ik}+c))/2)$) (right)  -- F-score, averaged over 30 runs when normalized Laplacian based GSC was applied.}
\label{tab:Data_0.N.Exp_5}
\label{tab:Data_0.N.Exp_4}
\begin{tabular}{|l|c|r|r|}
\multicolumn{3}{c}{ using formula \eqref{eq:sp_cos2c}  }\\
\hline
Embedding   &$c$& avg.   & SD of   \\
 type & &  F-score       &   F-score \\
\hline
CountVectorizer    &0  & 0.036 & 0.020  \\
&1&0.092&0.043\\
&2&0.088&0.028\\
&3&0.093&0.040\\
\hline
TfVectorizer         &0  & 0.035 & 0.017  \\
&1&0.077&0.043\\
&2&0.096&0.039\\
&3&0.088&0.044\\
\hline
TfidfVectorizer      &0  & 0.025 & 0.001  \\
&1&0.081&0.034\\
&2&0.073&0.036\\
&3&0.087&0.058\\
\hline
\hline
WikiGloVe     &0  & --- & ---  \\
&1 & 0.068 & 0.037  \\
&2&0.085&0.048\\
&3&0.075&0.044\\
\hline
TweetGloVe     &0  & 0.081 & 0.044  \\
&1&0.081&0.066\\
&2&0.081&0.054\\
&3&0.084&0.043\\
\hline  
\end{tabular}
\begin{tabular}{|l|c|r|r|}
\multicolumn{3}{c}{ using formula \eqref{eq:exp_c}  }\\
\hline
Embedding   &$c$& avg.   & SD of   \\
 type & &  F-score       &   F-score \\
\hline
CountVectorizer    &0  & 0.089 & 0.045  \\
&1&0.084&0.036\\
&2&0.102&0.045\\
&3&0.080&0.035\\
\hline
TfVectorizer         &0  & 0.085 & 0.031  \\
&1&0.084&0.037\\
&2&0.086&0.037\\
&3&0.109&0.048\\
\hline
TfidfVectorizer      &0  & 0.090 & 0.046  \\
&1&0.088&0.056\\
&2&0.082&0.038\\
&3&0.100&0.065\\
\hline
\hline
WikiGloVe     &0  & 0.082 & 0.047  \\
&1&0.093&0.053\\
&2&0.077&0.044\\
&3&0.079&0.046\\
\hline
TweetGloVe     &0  & 0.074 & 0.049  \\
&1&0.085&0.047\\
&2&0.088&0.047\\
&3&0.080&0.050\\
\hline  
\end{tabular}
\end{table}

Table \ref{tab:Data_0.N.Exp_1} {(left)} presents clustering results when the correction of negative similarities is based on adding a constant to all off-diagonal similarities. 
Normalized Laplacian-based clustering could be executed except for $c=0$ in WikiGlove embedding because the diagonal of $D$ contained negative entries. 
We see that modern (GloVe based) does not have any real advantage over traditional (TVS based) embeddings. 
At the same time, we see that adding the constant $c=1$ significantly improves the performance, while higher constants do not contribute much to the results. 

Table \ref{tab:Data_0.N.Exp_2} {(right)} presents clustering results when the correction of negative similarities is based on adding a constant to all off-diagonal similarities and dividing for normalization.  As expected, no improvement over Table \ref{tab:Data_0.N.Exp_1} {(left)} is visible. 

Table \ref{tab:Data_0.N.Exp_3}  (left) presents the results when normalizing over the largest angle between document vectors. The results are worse for TVS embeddings, and slightly worse for GloVe embeddings.

Table \ref{tab:Data_0.N.Exp_4}  (left) presents the results when dividing the angle between document vectors. The results constitute an improvement when dividing by at least two, but dividing by higher values does not contribute anything. 

Table \ref{tab:Data_0.N.Exp_5}  (right) presents the results when replacing primary similarities with their exponential variants.  
The variants do not differ much, but replacement of negative similarities with exponential ones helps the GloVe based embeddings, and also the TVS embeddings benefit from this transformation. 

As visible in the right tables  \ref{tab:Data_0.N.Exp_0}, \ref{tab:Data_0.N.Exp_3}, the results for combinatorial Laplacians are significantly worse, and the effects of transformations are generally marginal, as expected. 

Detailed results for all samples are available at \naszLink. Conclusions: removal of negative similarities allows to compute or improve clustering results for GloVe embeddings whereby setting negatives to zero is the worst method. The transformations studied are also helpful for TVS based clustrings, whereby the reason for this effect needs to be studied deeper.

\section{Conclusions}\label{sec:conclusions}

In this paper we discussed the issues in graph spectral clustering of documents resulting from growing popularity of embeddings different from the traditional Term Vector Space. The major problem is the negative cosine similarities between documents under these embeddings. 

The major known issue is that it may be impossible to compute normalized Laplacian because of negativity of the diagonal matrix $D$ entries. But we have shown experimentally that even if $D$ has a positive diagonal, negative similarities deteriorate clustering results. 

We have studied six different methods for overcoming negative similarities. 
Essentially, the combinatorial Laplacian-based clusterings seem to be unaffected by negative similarities, as demonstrated by theoretical arguments.  
In case of normalized Laplacians, the method of setting negative similarities to zero yields the worst results. The other methods perform similarly.
Interestingly, it turns out that for Term Vector Space embeddings there may be an improvement of performance when the similarity correction is applied.

We were also able to provide a geometric interpretation of one of the studied methods.  

This study was limited to two GloVe type embeddings, based on Wiki training data and Tweeter training data. 
Further investigation into other methods, such as BERT or other transformer-based methods, would be needed to make a more in-depth judgment. The study was performed on short documents, and verification of the results for longer ones may also be insightful. 
\bibliographystyle{splncs04}

 \end{document}